\documentclass[letterpaper, 10 pt, conference]{ieeeconf}  
\usepackage{xcolor}
\usepackage{graphicx}
\usepackage{booktabs}
\usepackage{amsmath}
\usepackage{newpxtext}

\IEEEoverridecommandlockouts                              

\title{\LARGE \bf
ActiveRIR: Active Audio-Visual Exploration for\\Acoustic Environment Modeling
}

\author{Arjun Somayazulu$^{1}$, Sagnik Majumder$^{1,2}$, Changan Chen$^{1}$, Kristen Grauman$^{1,2}$
\thanks{$^{1}$The University of Texas at Austin}%
\thanks{$^{2}$FAIR, Meta}%
}

\makeatletter
\newcommand{\ie}{\emph{i.e. }\@ifnextchar.{\!\@gobble}{}}
\newcommand{\cf}{\emph{c.f. }\@ifnextchar.{\!\@gobble}{}}
\newcommand{\eg}{\emph{e.g. }\@ifnextchar.{\!\@gobble}{}}
\newcommand{\etc}{etc\@ifnextchar.{}{.\@}}
\makeatother

\begin{document}

\maketitle
\thispagestyle{empty}
\pagestyle{empty}

\begin{abstract}
An \textit{environment acoustic model} represents how sound is transformed by the physical characteristics of an indoor environment, for any given source/receiver location. Traditional methods for constructing acoustic models involve expensive and time-consuming collection of large quantities of acoustic data at dense spatial locations in the space, or rely on privileged knowledge of scene geometry to intelligently select acoustic data sampling locations. We propose \textit{active acoustic sampling}, a new task for efficiently building an environment acoustic model of an unmapped environment in which a mobile agent equipped with visual and acoustic sensors jointly constructs the environment acoustic model and the occupancy map on-the-fly. We introduce ActiveRIR, a reinforcement learning (RL) policy that leverages information from audio-visual sensor streams to guide agent navigation and determine optimal acoustic data sampling positions, yielding a high quality acoustic model of the environment from a minimal set of acoustic samples. We train our policy with a novel RL reward based on information gain in the environment acoustic model. Evaluating on diverse unseen indoor environments from a state-of-the-art acoustic simulation platform, 
ActiveRIR outperforms an array of methods---both traditional navigation agents based on spatial novelty and visual exploration as well as existing state-of-the-art methods.
\end{abstract}

\section{INTRODUCTION}

The acoustic properties of the sounds a mobile agent hears are determined by the physical characteristics of the space it is in, such as the room's geometry, the objects within it, the types of materials that comprise the room and objects' surfaces, and the agent's 
proximity and orientation with respect to the sound source. Consider a conversation with someone standing across a large auditorium or gym, compared to the same conversation sitting a few feet apart in a small, carpeted living room; Large spaces with hard surfaces (\eg concrete, glass) will add reverberation to audio, while small, cluttered spaces covered in soft materials (\eg curtains, carpet) absorb sound waves quicker, producing dull and anechoic audio. These physical properties transform any sound emitted from a source location in the environment according to various acoustic phenomena, including direct sounds, early reflections, and late reverberations, before it reaches our ears. Together, these phenomena form a Room Impulse Response (RIR), which characterizes the transfer function between a sound emitted at the source location and the sound that reaches a microphone or our ears~\cite{62e5b0bcac3243f0b77291509411ed52}.

\begin{figure*}[ht]
  \centering
  \includegraphics[scale=0.25]{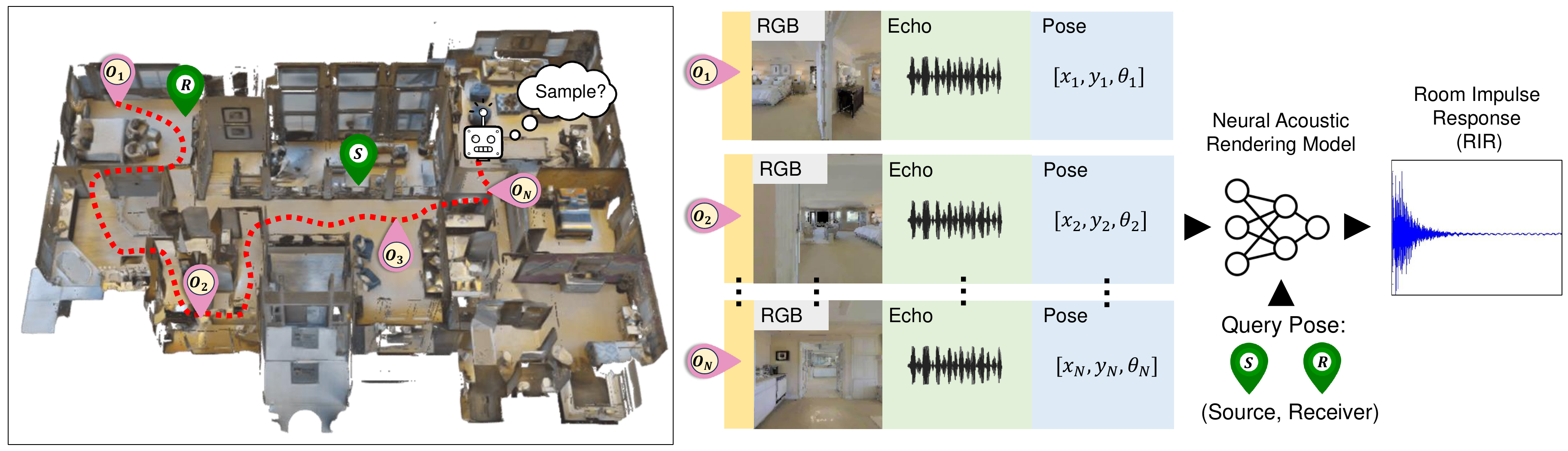}
  \caption{\textbf{Active acoustic sampling.} An agent must intelligently navigate an unmapped 3D scene and actively sample audio-visual observations (the scene's \textit{acoustic context}) to construct an acoustic model of the environment, given limited navigation time and a fixed sampling budget. When queried with an arbitrary sound source position and receiver pose in the space, the learned environment acoustic model should accurately generate the corresponding Room Impulse Response (RIR) at that pose.}
  \label{fig:concept_fig}
  \vspace{-0.1in}
\end{figure*}
    
An \textit{environment acoustic model} is a complete representation of a scene's acoustics~\cite{majumder2022fewshot, luo2023learning, ratnarajah2022mesh2ir, ratnarajah2024listen2scene, 9747846}.  Given a sound source location and a listener's position and orientation (pose) as a query, the model renders the corresponding RIR, accounting for all major acoustic phenomena due to the physical properties of the space.

Environment acoustic models are critical for robotics. In mobile robotics, an environment acoustic model can provide rich contextual information to an agent. Tasks such as navigating to a target sound~\cite{chen2021learning} and separating out a sound of interest from background sounds~\cite{majumder2022active, majumder2021move2hear} require an agent to decide where to move based on the audio it hears. An agent equipped with the environment acoustic model can better anticipate the effects of its movement on the observed audio.
In AR/MR applications, acoustic models  allow a virtual sound source (such as a human speaker)
inserted at a position in the user's real-world space to sound properly spatialized with acoustics that match the space, as the user moves around their environment.

Capturing an RIR is a physically involved process. A loudspeaker must be placed at the desired height and location, and a microphone setup placed at a similar height at the desired receiver location and orientation. The speaker emits a sound impulse, and the receiver microphone(s) record the resulting RIR, which can last several seconds. 

Existing methods for environment acoustic modeling assume extensive physical access to the environment in order to collect these RIRs at arbitrary positions.
Neural Field methods~\cite{luo2023learning, liang2023neural} require large quantities of RIRs captured at dense source/listener locations (approx. every 1 meter) throughout the environment, which can be expensive and time-consuming. A few-shot approach~\cite{majumder2022fewshot} overcomes the intensive data requirement, building acoustic models of novel environments from a limited number of observations, but still requires extensive knowledge of the floorplan in order to sample uniformly spaced locations around obstacles. Other methods rely heavily on knowledge of the scene geometry from 3D meshes~\cite{ratnarajah2022mesh2ir} or floorplans~\cite{9747846}. Importantly, these methods all assume \emph{instantaneous access to observations at arbitrary positions in the environment}, which is unrealistic in the embodied robotics context---where an agent must physically travel between locations---and in unmapped environments where no prior knowledge of the floorplan and obstacles is available.

We introduce \emph{active acoustic sampling}, a new task that requires a single mobile agent with audio-visual sensing to efficiently construct an unmapped environment's acoustic model within a total budget of acoustic samples, despite only on-the-fly discovery of its floorplan. The proposed problem is distinct from traditional visual exploration~\cite{ramakrishnan2020exploration, NIPS2016_afda3322, strehl_analysis_2008}, where agents prioritize motions in a scene to rapidly complete the occupancy map. While in floorplan mapping the best places to reach are those that add visibility to the widest floor area, in acoustic modeling the most valuable poses (sampling spots) in the environment depend on all aspects of the 3D geometry and surface materials. 

To address this challenge, we present ActiveRIR, an active sampling policy that can be deployed on mobile agents in environments that are both \textit{unseen} and \textit{unmapped}. ActiveRIR is trained with a novel audio-visual exploration reward to guide wide agent exploration and inform the agent's decision on when to sample an acoustic observation, within a budget of total acoustic samples. Our acoustic reward measures the global improvement in the agent's environment acoustic model estimate after an observation is sampled, ensuring that the limited context used to build the environment acoustic model contains only the most valuable observations seen by the agent during its exploration. 

Evaluating on a diverse set of unseen and unmapped
scanned real-world 3D indoor environments together with state-of-the-art (SOTA) acoustic~\cite{chen2023soundspaces} and visual~\cite{Matterport3D} scene simulation platforms, ActiveRIR produces a higher-quality acoustic model in >70\% fewer steps than passive approaches, and outperforms both traditional visual and spatial exploration-based methods~\cite{Savinov2019_EC, ramakrishnan2020exploration, strehl_analysis_2008, NIPS2016_afda3322} as well as SOTA scene acoustic modeling methods~\cite{majumder2022fewshot}. We also demonstrate that the performance gain achieved using ActiveRIR-collected observations generalizes across multiple acoustic rendering methods, showing promising potential for ActiveRIR to be plugged as a module into existing acoustic rendering methods and improve the quality of generated environment acoustic models.

\section{RELATED WORK}
\subsection{Environment acoustics and mapping}
Estimating room acoustics from images of indoor scenes has been widely explored, including explicit RIR estimation from images~\cite{singh2021image2reverb, RemaggiLuca2019RRWA} as well as methods that model its acoustic transformation on source sounds or speech~\cite{chen2022visual, gao201925d, somayazulu2023selfsupervised, xu2021visually, 9711121}. While these methods can produce audio that perceptually matches the general acoustics of the space (\eg a concert hall vs. bedroom), they cannot reason about fine-grained acoustic effects of shifts in speaker or listener pose \textit{within} a visual scene. 

Audio field coding approaches~\cite{raghuvanshi_parametric_2014, 6777442, 10.1145/3386569.3392459} estimate generic perceptual RIR features using parametric sound field representations that model spatial relationships between acoustics at different positions. Recent approaches use Neural Fields to generate RIRs~\cite{luo2023learning} directly, though they still require large-scale acoustic data (>10k samples) from dense spatial locations, and the learned model cannot generalize to other environments.

A transformer-based approach produces an acoustic model of an unseen environment given limited acoustic data sampled randomly from the floorplan~\cite{majumder2022fewshot}. Other GNN-based approaches generate RIRs given a full 3D scene mesh and dataset of acoustic material coefficients~\cite{ratnarajah2022mesh2ir, ratnarajah2024listen2scene}. These methods rely on prior knowledge of scene geometry, and assume the ability to instantaneously access visual and acoustic observations at selected positions in the space. Access to this privileged information---as well as the ability to teleport to new locations without penalty---are significant assumptions that are unrealistic for robots. In principle, the process that produced a pre-computed floorplan map or mesh could have also pre-computed the acoustic model with densely sampled observations using minimal additional time or energy. In short, assuming access to full scene geometry but \emph{not} complete RIRs  as well fails to capture real-world constraints in robotics.

\vspace{-0.03in}
\subsection{Audio-visual navigation and exploration}

Equipping mobile agents with sound production and audio capture has led to a proliferation in audio-visual embodied tasks, such as audio-visual sound source separation~\cite{majumder2021move2hear, majumder2022active}, audio-visual navigation~\cite{chen2020soundspaces,chen2021learning,chen2021semantic,yu2022sound}
and audio-visual floorplan mapping~\cite{purushwalkam2020audio, majumder2023chat2map}. A mobile audio-visual agent can decide where to emit and receive acoustic observations to help with floorplan mapping~\cite{hu2023active}. While mapping the floorplan requires dense exploration of the space and its frontiers,
we hypothesize that an accurate environment acoustic model of the same space can be built with far fewer acoustic observations sampled intelligently at select locations.
Prior work~\cite{yu_measuring_2023} trains a policy to construct an environment acoustic model that relies on two mobile agents: an "emitter" emits an impulse sound and the "receiver" records the RIR. The agents navigate according to a local acoustic reward that measures the predicted RIR error at the agents' next position. However, unlike our global formulation, movement to minimize local acoustic prediction accuracy often prevents the agent from navigating towards areas with challenging and dynamic acoustics (\eg hallways with turns and corners), hurting performance  (see Sec.~\ref{sec:experiments}).

\section{Active acoustic sampling task}\label{sec:task}

We introduce the task of \textit{active acoustic sampling}. The goal of this task is to train a mobile agent to navigate an unmapped 3D environment (such as a home or office) and intelligently sample egocentric audio-visual observations ("acoustic context") within a predefined \emph{sample budget} and \emph{time budget}, such that a acoustic rendering model conditioned on this context can produce accurate and high-fidelity RIRs given arbitrary query source/receiver poses. Each audio-visual observation consists of the agent's egocentric RGB-D image and an echo response, namely an RIR obtained 
by placing the sound source and receiver microphones at the location of image capture, emitting a sinusoidal frequency sweep signal and recording the echoes~\cite{majumder2022fewshot}. See Fig.~\ref{fig:concept_fig}.
Our sample budget is much lower than the timestep budget for navigation, requiring the agent to carry out the memory-intensive~\cite{yu_measuring_2023} task of capturing and storing echo responses only at a select few locations most valuable for the global scene acoustic model. The audio sampling is preemptive, \ie the model decides to choose or skip an audio sample before capturing it.

Formally, given navigation time budget $T$ and audio sample budget $N \ll T$, the agent must navigate an unmapped scene and collect audio-visual samples $\mathcal{C} = \{C_i\}^{N}$, where $C_i = (A_i, V_i, P_i)$. $A_i$ denotes the binaural (two-channel) echo response, and $V_i$ is the $90^\circ$ FoV RGB-D image captured at the camera pose $P_i = (x_i, y_i, \theta_i)$ at location $(x_i, y_i)$ and orientation $\theta_i$.

Using $\mathcal{C}$ as the context, the agent must infer the scene's environment acoustic model, such that it can accurately estimate the RIR $R^Q$ for arbitrary query $Q=(s_j, l_k)$, where $s_j = (x_j, y_j)$ is an omni-directional sound source at location $(x_j, y_j)$, and $l_k=(x_k, y_k, \theta_k)$ is the receiver microphone pose. Thus, the task objective is to learn a policy $\pi$ that guides our agent towards acoustically informative locations in the scene and helps it decide where to sample audio-visual observations to add to the acoustic context $\mathcal{C}$, such that a acoustic rendering model $f$ conditioned on $\mathcal{C}$ approximates the ground-truth RIR $R^Q$ for an arbitrary query $Q$, or $\tilde{R}^Q = f(Q | \mathcal{C})$. At each time step $t$, where $1\leq t\leq T$, the policy can take action $\alpha_t \in \mathcal{A}$, where $\mathcal{A}=\big\{MoveForward, TurnLeft, TurnRight\big\} \times \big\{Sample,Skip\big\}$ is the agent's action space.
    
This task relies on context cues from both audio and vision. RGB-D images capture local room geometry, the presence of furniture/obstacles, and the materials of surfaces in view, while acoustic observations help the agent associate these physical properties with their acoustic effects on an emitted sound. Audio also captures longer-range room geometry beyond the agent's current FoV, which vision cannot. Exploiting this visual-acoustic correspondence is key in helping the agent decide how to move and where to \emph{Sample} an acoustic observation.

\begin{figure*}
  \centering
  \includegraphics[width=0.85\linewidth]{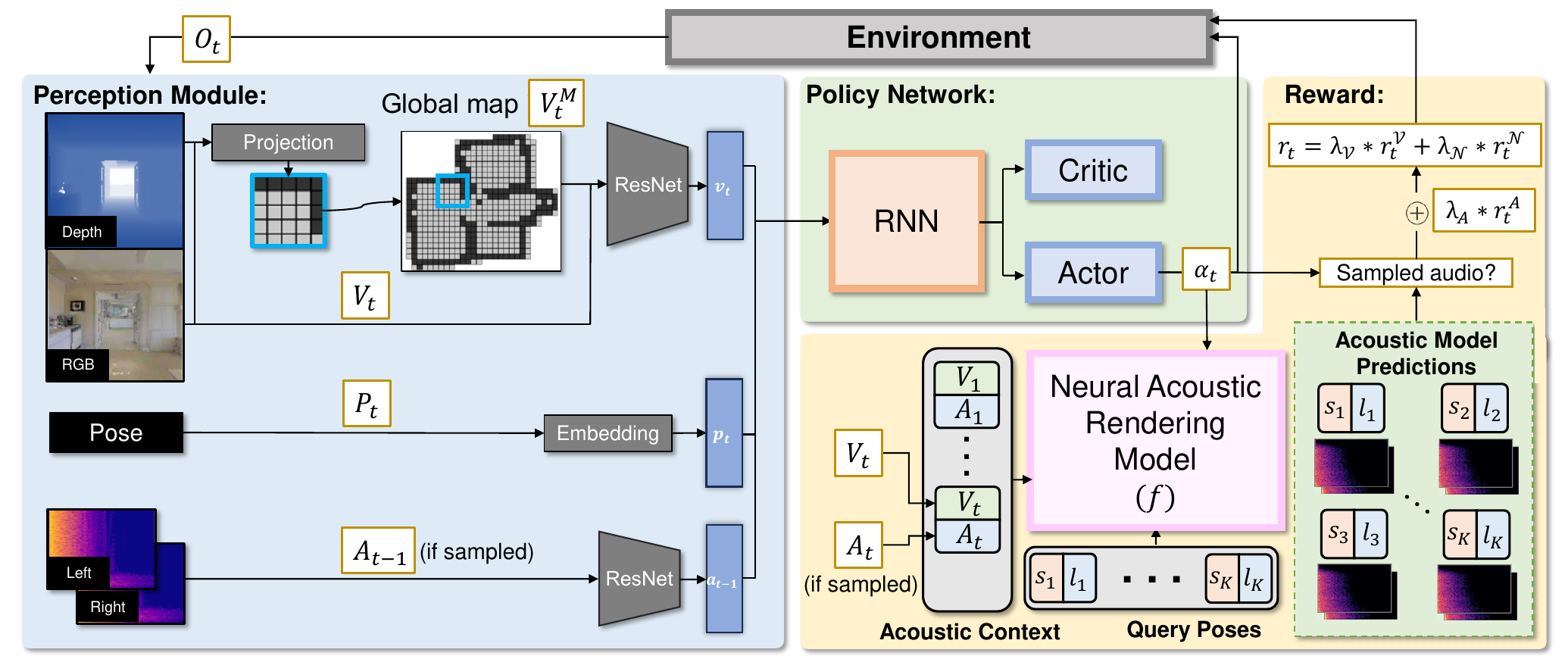}
  \vspace{-0.1in}
  \caption{\textbf{ActiveRIR policy network architecture and reward.} At each step $t$, our policy $\pi$ receives an egocentric visual input $V_t$, the camera pose $P_t$, and the binaural echo response $A_{t-1}$---if it was sampled by the policy at the previous step---and predicts an action $\alpha_t$ that decides both how the agent should move, and if it should sample the current echo response $A_t$. It then uses $A_t$ along with the current visual input $V_t$ to improve its acoustics prediction quality. Given these audio-visual samples ("Context") collected over an episode, the agent uses an off-the-shelf acoustic rendering model to predict the RIR for any arbitrary query pair of sound source and receiver locations. We train our policy with an audio-visual reward which encourages healthy exploration of the scene in search of acoustically important locations, and guides the agent \textit{when} to sample highly valuable observations, subject to a maximum audio sample budget.}
  \vspace{-0.15in}
  \label{fig:model_fig}
\end{figure*}
\section{Approach}

We pose the task as a reinforcement learning problem, where a model sequentially decides where to sample audio-visual observations given visual RGB-D frames and previously sampled audio\footnote{All audio inputs to our model are represented as two-channel magnitude spectrograms computed using the short-time Fourier transform (STFT)~\cite{majumder2022fewshot}.}, and the environment acoustic model is built from this collected context. We propose \textbf{ActiveRIR}, which consists of \textbf{1)} an audio-visual sampling policy $\pi$, and \textbf{2)} a neural acoustic rendering model $f$ for predicting scene acoustics using the samples. See Fig.~\ref{fig:model_fig}. Next, we describe these components in detail.

\vspace{-0.05in}
\subsection{Audio-visual sampling policy}
\label{sec:active_sampling}
\subsubsection{Policy inputs and architecture}
At every time step $t$, our audio-visual sampling policy $\pi$ receives $O_t$ as the input and decides how to move in the environment, and whether to capture and add the echo response at the current step to our acoustic context (i.e., \emph{Sample} or \emph{Skip}). Formally, $O_t=(A_{t-1}, V_t, P_t)$, if the echo response was sampled at the previous step, and $O_t=(V_t, P_t)$ if it was skipped. Before encoding $O_t$, $\pi$ pre-processes $V_t$ to generate $V^{\pi}_t=(V^\mathcal{R}_t, V^{M}_t)$, where $V^\mathcal{R}_t$ is the RGB image from $V_t$, and $V^{M}_t$ is a topdown global occupancy\footnote{$V^{M}_t$ is produced by projecting the current and past depth images $V^D_{1\ldots t}$ to the egocentric ground plane and stitching together these projections into a 
cumulative map of the scene.} map. Note that the map begins empty and accumulates as the agent selects its motions.

First, we feed $A_{t-1}$ (if sampled), $V^{\pi}_t$ and $P_t$ to separate encoders and extract audio features $a_{t-1}$, visual features $v_t$ and pose features $p_t$.
The visual and audio encoders are ResNets~\cite{he2016deep}, and the pose encoder is a sinusoidal positional embedding~\cite{vaswani2017attention}. Next, we concatenate $a_{t-1}$, $v_t$ and $p_t$ into $o_t$ and feed it to the policy network, which consists of an RNN and an actor-critic module. The RNN estimates an updated history $h_t$ along with the current state representation $g_t$, using the fused feature $o_t$ and the history of states $h_{t-1}$. The actor-critic module takes $g_t$ and $h_{t-1}$ as inputs and predicts a policy distribution $\pi_{\phi}(\alpha_t|g,t, h_{t-1})$ along with the value of the state $H_{\phi}(g_t, h_{t-1})$. Finally, the policy samples an action $\alpha_t$ from its action space $\mathcal{A}$ (\cf Sec.~\ref{sec:task}) per the distribution $\pi_{\phi}$.

\subsubsection{Policy reward}
\label{sec:policy_reward}
We propose a novel RL reward to train our policy $\pi$:\
\begin{equation}
    r_t = \lambda_A * r^A_t + \lambda_\mathcal{V} * r^\mathcal{V}_t + \lambda_\mathcal{N} * r^{\mathcal{N}}_t.
\end{equation}
Here $r^A_t$ is our novel \textit{Acoustic Prediction} reward, which measures improvement in the environment acoustic model from the previous step $t-1$ to the current step $t$ if audio was sampled, and is given by 

$$r^A_t = \mathcal{L}^R_{t-1} - \mathcal{L}^R_{t}$$

, where $\mathcal{L}^R_t$ is the mean L1 distance between the predicted and ground-truth RIR magnitude spectrograms for a fixed set of $K$ query positions selected from throughout the (training) scene, at step $t$. $r^\mathcal{A}_t$ is zero at the steps where the policy decides not to sample; thus it encourages the agent to sample an acoustic observation \emph{when the agent is at a position that will improve the global acoustic prediction quality.}

Since this sparse reward can impact the stability of RL training, we augment $r^\mathcal{A}_t$ with an area-coverage reward~\cite{chen2019learning} 
$r^\mathcal{V}_t = (\mathcal{V}_t - \mathcal{V}_{t-1})/\mathcal{V}_{t-1}$ that measures the relative increase in area coverage over time, 
where $\mathcal{V}_t$ and $\mathcal{V}_{t-1}$
are the total area covered by the agent at steps $t$ and $t-1$, respectively.
We also add a novelty reward~\cite{NIPS2016_afda3322, Savinov2019_EC, ramakrishnan2020exploration} $r^{\mathcal{N}}_{t} = \frac{1}{\sqrt{n(m_t)}}$, where $n(m_t)$ is the visitation count at 1$\times$1 meter discretized floorplan grid cell $m_t$. $\lambda_\mathcal{V}$ and $\lambda_\mathcal{N}$
are the respective reward weights. Whereas $r^{\mathcal{V}}_t$ rewards the agent for taking actions that expose new areas of the scene to the agent, $r^{\mathcal{N}}_t$ incentivizes increasing the visitation count of novel locations in the scene. $r^{\mathcal{V}}_t$ and $r^{\mathcal{N}}_t$ together promote a healthy exploration of the scene, which is crucial for learning a good policy, and also encourage stable RL training owing to their dense nature. We train $\pi$ using Decentralized Distributed PPO (DD-PPO)~\cite{wijmans2019dd}. The DD-PPO loss consists of a value loss, policy loss, and an entropy loss for further improving exploration.

\begin{figure*}[ht]
  \centering
  \includegraphics[width=0.97\linewidth]{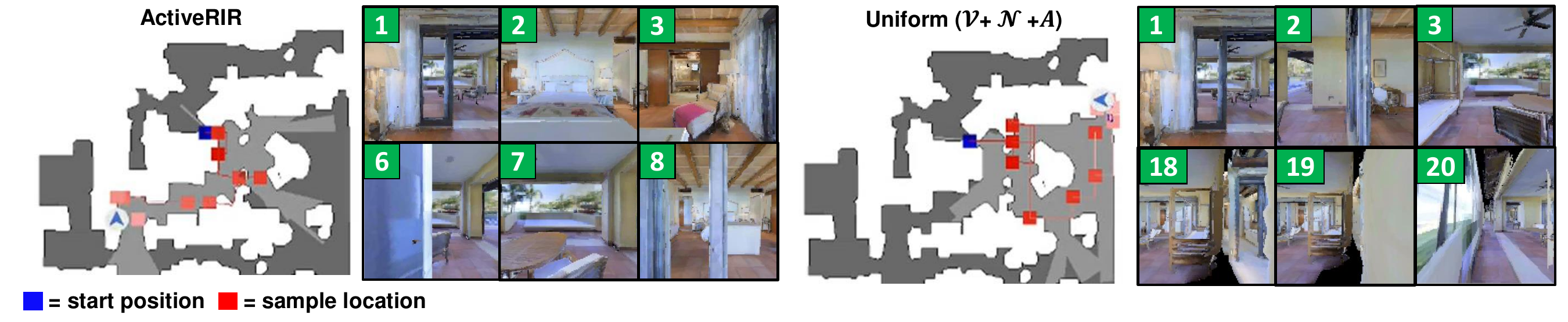}
  \caption{\textbf{ActiveRIR vs. Uniform sampling.} The ActiveRIR agent (\textbf{far left}) navigates an environment and actively samples observations, collecting context \textbf{(left)} from regions of the environment where acoustics rapidly change---such as in a winding hallway---and which are visually \textit{and} acoustically distinct from other samples in the context. In contrast, an agent passively sampling at a uniform interval \textbf{(right)} collects an acoustic context \textbf{(far right)} with spatial and visual redundancy, as observed by the bottom two images which show similar views of the same room captured only 1 meter apart.}
  \label{fig:full_vis_fig}
  \vspace{-4mm}
\end{figure*}

\vspace{-0.04in}
\subsection{Acoustic rendering model}
\label{sec:rir_pred_network}
Our sampling policy design is agnostic of the design of the acoustic rendering model $f$, providing the flexibility of using our policy with any off-the-shelf rendering model that can take audio-visual samples as inputs and predict the RIR $R^Q$ at a source-receiver query $Q$ in the scene (\cf Sec.~\ref{sec:task}). We use FewShot-RIR (FS-RIR)~\cite{majumder2022fewshot} as our rendering model backbone, due to its state-of-the-art performance. We also evaluate ActiveRIR's ability to generalize in experiments with a second backbone, NAF~\cite{luo2023learning} (\cf Sec.~\ref{fig:naf_fsrir}). Given audio-visual samples from a scene, FS-RIR uses a transformer~\cite{vaswani2017attention} encoder to build an acoustic context of the scene, followd by a transformer-decoder to predict the RIR for an arbitrary source-receiver query using the acoustic context. We pre-train the acoustic rendering model $f$ in a disembodied fashion---without a sampling policy---randomly selecting observations from a scene to add to the acoustic context. Next, we train our policy with RL while keeping $f$ frozen. This helps improve RL training stability by ensuring stationarity in the reward distribution (\cf Sec.~\ref{sec:active_sampling}).
\vspace{-0.1in}

\section{Experiments}
\label{sec:experiments}

\vspace{-0.05in}

\subsection{Experimental setup}
We use the state-of-the-art SoundSpaces (SS) 2.0~\cite{chen2023soundspaces} acoustics simulator, built on top of the AI-Habitat~\cite{savva2019habitat} simulator and Matterport3D (MP3D)~\cite{Matterport3D} scenes. MP3D consists of photorealistic scans of diverse, real-world multi-room indoor environments complete with furniture and other objects (\eg tables/sofas/desks/lamps). 
 SS 2.0 supports continuous rendering of precise spatial audio in arbitrary 3D scenes, capturing all major acoustic phenomena~\cite{chen2023soundspaces}. We use 78 diverse MP3D scenes, split into train/val/test sets of 56/10/12, preserving diverse scene types and sizes within each split. We train our policy for 150K PPO iterations and validate/test on 123/225 inference episodes respectively sampled in proportion to scene sizes within each split.

We place the agent at a random location in a scene at the start of every episode. We set $T=200$ to account for the minimal time needed to traverse an average scene average in Matterport3D~\cite{Matterport3D}, and
set the audio sample budget to $N=20$ samples to match the context size used in FS-RIR~\cite{majumder2022fewshot}. We set $\lambda_{A} =2\times10^5$,
$\lambda_{\mathcal{V}} =2\times10^2$
and $\lambda_{\mathcal{N}} = $10.0 to place the component rewards on the same scale. The agent has turn and movement resolution 90$^\circ$ and 1 meter respectively. We evaluate performance with \textbf{STFT L1 Error (STFT)}~\cite{majumder2022fewshot, chen2022visual}, which measures mean L1 error between 
predicted and ground-truth RIR magnitude spectrograms at
$K=60$ global query poses per scene, sampled randomly at the scene level. 

\vspace{-0.03in}
\subsection{Baselines}\label{sec:baselines}
\textbf{Random agent:} an agent that chooses an action from our action space $\mathcal{A}$ randomly.
\textbf{Forward agent}: 
an agent that only moves forward, sampling observations uniformly (every $\frac{T}{N}$ steps). \textbf{Greedy agent:} an agent that navigates randomly, and greedily selects the first $N$ observations on its path.

\section{Results}

\begin{figure}
  \centering
  \includegraphics[scale=0.33]{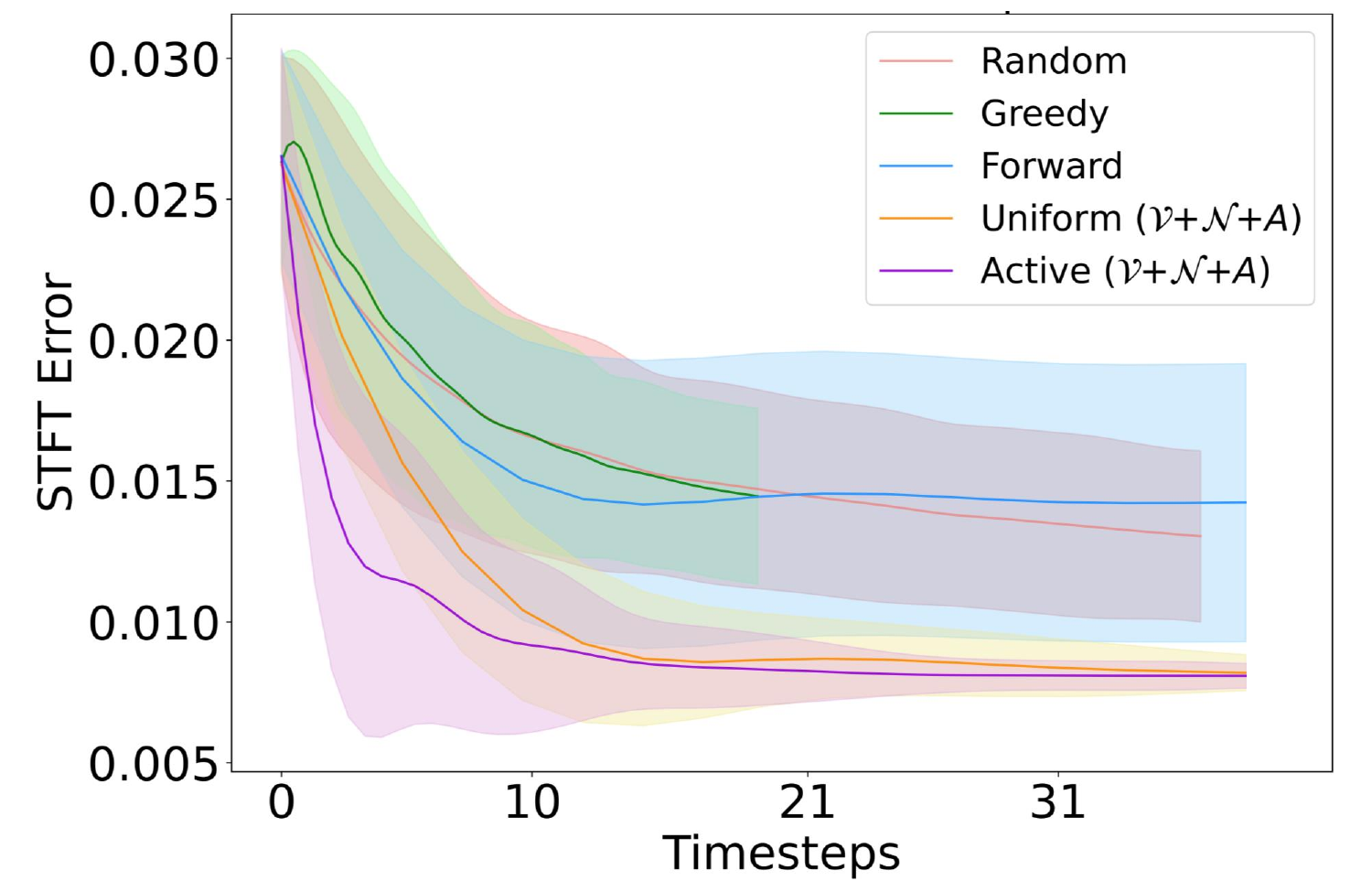}
  \vspace{-3mm}
  \caption{\textbf{Acoustic prediction quality vs timesteps.} ActiveRIR (purple) rapidly minimizes STFT error in the acoustic model in fewer steps than an acoustic agent sampling at a fixed interval (orange), and outperforms heuristic approaches as well.}
  \label{fig:active_vs_uniform}
  \vspace{-5mm}
\end{figure}

Table~\ref{table:main} shows acoustic prediction performance on unseen test environments. ActiveRIR significantly outperforms naive approaches (Greedy, Forward, Random). Fig.~\ref{fig:active_vs_uniform} displays STFT error as a function of the agent's steps. ActiveRIR (purple) efficiently navigates toward and captures valuable acoustic observations that rapidly minimize STFT error in the acoustic model, compared both to a passive-sampling audio-visual agent which samples every $\frac{T}{N}$ steps ("Uniform", orange) as well as an array of heuristic approaches.

\vspace{0.1in}
\subsubsection{Model analysis}
\paragraph{Reward variations}
To evaluate the impact of our Acoustic Prediction reward, we train active policies with ablations of our audio-visual reward (Table~\ref{table:ablations}). We outperform Coverage ($\mathcal{V}$) and Novelty ($\mathcal{N}$) agents as well as the Exploration agent ($\mathcal{V}$+$\mathcal{N}$), which was trained with only the spatio-visual component of our audio-visual reward (\cf Sec.~\ref{sec:policy_reward}), validating the importance of acoustic information in the agent's decision to \textit{Sample} an observation beyond what can be inferred from visual and/or spatial sensory information alone. We also significantly outperform a passive audio-visual agent (Uniform), confirming that actively determining \textit{when} to sample plays a critical role in collecting valuable acoustic context beyond simply navigating \textit{towards} acoustically and visually novel areas. Inspired by~\cite{yu_measuring_2023}, we evaluate a local variant of our acoustic reward, defined as the improvement in the RIR prediction error for a query closest to the agent's current position. We outperform this local acoustic agent, demonstrating that our global acoustic reward helps collect samples most valuable for \textit{global} acoustic model.

\paragraph{Pose sensor noise}
We evaluate ActiveRIR's robustness to Gaussian noise in pose and actuation. STFT error only increases from $8.08 \times 10^{-3}$ to $8.10 \times 10^{-3}$, highlighting the effectiveness of our design choices.

\begin{figure}
  \centering
  \includegraphics[scale=0.31]{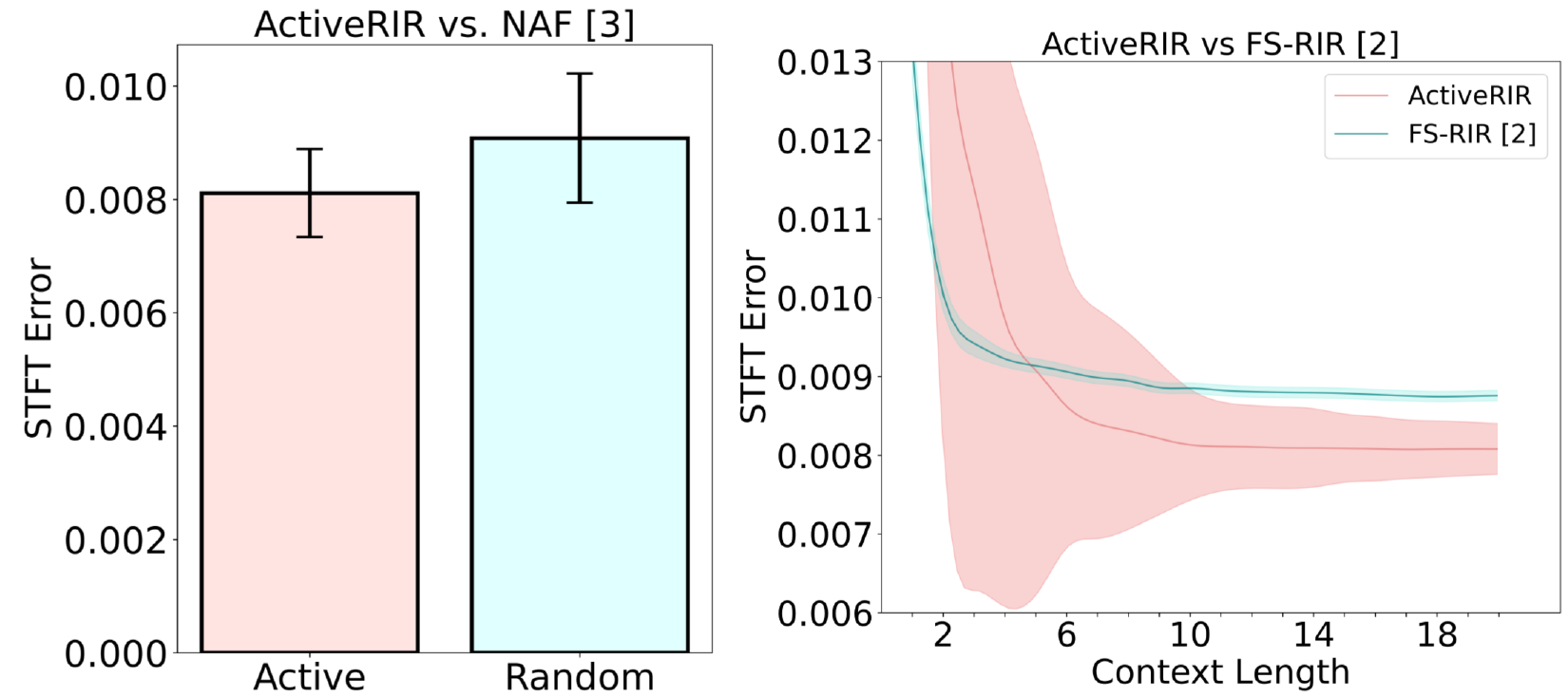}
  \vspace{-8mm}
  \caption{\textbf{Active sampling with existing methods.} \textbf{(Left)} NAF~\cite{luo2023learning} trained on ActiveRIR-collected context outperforms NAF trained on context collected by a random policy, demonstrating ActiveRIR's ability to select valuable acoustic context agnostic of the acoustic rendering model. \textbf{(Right)} As we grow the context size, ActiveRIR samples high-value observations that rapidly improve global scene acoustic error, producing a final acoustic model with significantly lower error than FS-RIR~\cite{majumder2022fewshot}.} 
  \vspace{-6mm}
  \label{fig:naf_fsrir}
\end{figure}

\subsubsection{Comparison with SOTA} Fig.~\ref{fig:naf_fsrir} compares ActiveRIR to SOTA methods without embodiment and navigation budget. First, we compare against FS-RIR~\cite{majumder2022fewshot}, which collects samples randomly throughout the environment (Fig.~\ref{fig:naf_fsrir}, right). Despite initial progress, FS-RIR plateaus as the episode progresses, while ActiveRIR continues selecting highly-informative samples that further reduce STFT error.
Importantly, ActiveRIR produces a final acoustic model with significantly lower error than FS-RIR, \textit{despite FS-RIR's prior knowledge of the floorplan for randomly sampling observations}. Furthermore, to achieve the error of ActiveRIR, FS-RIR needs to acquire a context length of 86 samples, more than 4x the samples required for our model. While randomly sampling from throughout the space may outperform our policy in high-resource regimes, ActiveRIR strongly outperforms FS-RIR in cost-efficient settings with restrictive time or sample budgets---which are practical considerations when considering physical constraints on access/time spent in the space, and the intrusive nature of emitting a sound impulse.

To determine the value of acoustic context collected by ActiveRIR agnostic of the acoustic rendering model $f$, we train two Neural Acoustic Field (NAF)~\cite{luo2023learning} models on the test scenes using 1) ActiveRIR-collected context and 2) context collected by a random policy, and evaluate mean STFT error across 100 query poses sampled randomly from throughout the scene (Fig~\ref{fig:naf_fsrir}, left). The ActiveRIR-trained NAF consistently outperforms NAF trained on the random policy's collected context, demonstrating that ActiveRIR collects a rich, generalizable acoustic context and is flexible with respect to the acoustic rendering model.

\subsubsection{Qualitative analysis}
Fig.~\ref{fig:full_vis_fig} visualizes the importance of active sampling. ActiveRIR 
captures audio-visual samples in areas where acoustics can dynamically change, such as winding hallways (far left), while also ensuring that the context is spatially and visually distinct, as can be observed by the diverse views in the active context (left). In contrast, a passive 
agent (right) does not actively avoid capturing redundant observations (bottom two images), and collects samples in the main cavity of the scene where acoustics are relatively stationary between nearby poses. Also see submitted video.

\begin{table}[t]
\vspace*{3mm}
\centering
\begin{tabular}{c|c}
\toprule
 Policy & STFT Error $\downarrow$\\
\midrule
 Greedy & 14.46\\
 Forward & 14.68 \\
 Random & 13.04\\
\midrule
 ActiveRIR ($\mathcal{V}$+$\mathcal{N}$+$A$) & \textbf{8.08} \\
\bottomrule
\end{tabular}
\caption{\textbf{Acoustic prediction performance.} Reported in base $10^{-3}$.}
\label{table:main}
\vspace{-6mm}
\end{table}

\begin{table}[t]
\centering
\begin{tabular}{c|c|c}
\toprule
 Reward & Sampling & STFT Error $\downarrow$\\
\midrule
  $\mathcal{V}$+$\mathcal{N}$+$A$ & Uniform & 8.24 \\
 $\mathcal{V}$ & Active & 8.15\\
 $\mathcal{N}$ & Active & 8.11 \\
 $\mathcal{V}$+$\mathcal{N}$ & Active & 8.15\\
 $\mathcal{V}$+$A$ & Active & 8.21 \\
 $\mathcal{N}$+$A$ & Active & 8.15 \\
 $\mathcal{V}$+$\mathcal{N}$+$A$ (local) & Active & 8.11 \\
 $\mathcal{V}$+$\mathcal{N}$+$A$ & Active & \textbf{8.08} \\
\bottomrule
\end{tabular}
\caption{\textbf{Reward variations in ActiveRIR.}
Reported in base $10^{-3}$.}
\label{table:ablations}
\vspace{-7mm}
\end{table}

\section{CONCLUSIONS}

We propose a new task, \textit{active acoustic sampling}, in which an agent must navigate an unmapped environment and collect audio-visual samples to construct a model of the scene's acoustics within a given time and sample budget. We propose an active sampling policy trained with a novel audio-visual exploration reward which guides an agent to navigate towards and select high-value audio-visual samples that yield a high-quality acoustic model. Our policy outperforms passive approaches in >70\% fewer timesteps, and outperforms robust visual and spatial exploration agents as well as SOTA environment acoustic modeling methods~\cite{majumder2022fewshot} across diverse indoor scenes. We show that the performance gain using ActiveRIR-collected samples generalizes across multiple acoustic rendering models, demonstrating promising potential for ActiveRIR to improve existing acoustic rendering methods. In future work, we plan to explore 3D scene reconstruction from acoustic exploration.

\clearpage
\bibliographystyle{IEEEtran}
\bibliography{IEEEabrv, IEEEexample, references}
\addtolength{\textheight}{-12cm}

\end{document}